%% file: main.tex
\documentclass[11pt]{article}

\usepackage[margin=1in]{geometry}
\usepackage{graphicx}
\usepackage{xcolor}
\usepackage{caption}
\usepackage{subcaption}
\usepackage{float}
\usepackage{amsmath}
\usepackage{amssymb}
\usepackage{listings}
\usepackage{url}
\usepackage{tcolorbox}
\usepackage{adjustbox}
\usepackage{algorithm}
\usepackage[noend]{algpseudocode}
\usepackage[numbers]{natbib}
\usepackage[hidelinks]{hyperref}

\title{\textbf{NCO4CVRP}: Neural Combinatorial Optimization for the Capacitated Vehicle Routing Problem}
\author{%
\begin{tabular}{c@{\hspace{2em}}c}
Mahir Labib Dihan$^{\dagger}$ & Md. Ashrafur Rahman Khan
\end{tabular}\\[0.75em]
\begin{tabular}{c@{\hspace{2em}}c@{\hspace{2em}}c}
Wasif Jalal & Md. Roqunuzzaman Sojib & Mashroor Hasan Bhuiyan
\end{tabular}\\[0.75em]
Department of Computer Science and Engineering\\
Bangladesh University of Engineering and Technology\\[0.5em]
\normalsize $^{\dagger}$Corresponding to \texttt{mahirlabibdihan@gmail.com}
}
\date{}

\newcommand{\chapter}[1]{\section{#1}}

\begin{document}

\maketitle

\begin{abstract}
Neural Combinatorial Optimization (NCO) has emerged as a powerful framework for solving combinatorial optimization problems by integrating deep learning-based models. This work focuses on improving existing inference techniques to enhance solution quality and generalization. Specifically, we modify the Random Re-Construct (RRC) approach of the Light Encoder Heavy Decoder (LEHD) model by incorporating Simulated Annealing (SA). Unlike the conventional RRC, which greedily replaces suboptimal segments, our SA-based modification introduces a probabilistic acceptance mechanism that allows the model to escape local optima and explore a more diverse solution space. Additionally, we enhance the Policy Optimization with Multiple Optima (POMO) approach by integrating Beam Search, enabling systematic exploration of multiple promising solutions while maintaining diversity in the search space. We further investigate different inference strategies, including Softmax Sampling, Greedy, Gumbel-Softmax, and Epsilon-Greedy, analyzing their impact on solution quality. Furthermore, we explore instance augmentation techniques, such as horizontal and vertical flipping and rotation-based augmentations, to improve model generalization across different CVRP instances. Our extensive experiments demonstrate that these modifications significantly reduce the optimality gap across various Capacitated Vehicle Routing Problem (CVRP) benchmarks, with Beam Search and SA-based RRC consistently yielding superior performance. By refining inference techniques and leveraging enhanced search strategies, our work contributes to the broader applicability of NCO models in real-world combinatorial optimization tasks.
\end{abstract}

\input{intro}
\input{exact}
\input{approximation}
\input{heuristics}

\input{metaheuristics}
\input{nco}

\clearpage
\bibliographystyle{plain}
\bibliography{references}
\end{document}

%% file: intro.tex
\section{Introduction} \label{sec:intro}


Combinatorial optimization problems are fundamental to numerous real-world applications, such as logistics, scheduling, and network design. These problems require finding an optimal solution from a finite set of possibilities, constrained by feasibility requirements. Traditional solution methods, including exact algorithms and heuristics, often struggle with scalability as problem sizes grow. This limitation has driven the development of innovative approaches like Neural Combinatorial Optimization (NCO), which leverages advances in machine learning to tackle these challenges.

NCO combines neural networks with algorithmic principles to learn solution strategies directly from data, bypassing the reliance on handcrafted heuristics. Using supervised or reinforcement learning, NCO models are trained to approximate optimal solutions or maximize problem-specific rewards. Reinforcement learning-based techniques, such as Policy Optimization with Multiple Optima (POMO), have demonstrated strong performance on routing problems, offering scalable and effective solutions.

The Capacitated Vehicle Routing Problem (CVRP) is a classic combinatorial optimization problem and an ideal candidate for exploring NCO. It involves designing routes for vehicles to service customers while minimizing costs, adhering to vehicle capacity limits, and ensuring each customer is served exactly once. CVRP's complexity and practical significance make it a valuable testbed for NCO methods, enabling insights into how these models address real-world optimization challenges.

This report focuses on applying NCO to the CVRP, exploring how neural networks can learn effective heuristics for solving this problem. The report is structured as follows: 

\begin{itemize}
    \item \textbf{Section 2} provides an overview of traditional exact algorithms used for solving combinatorial problems, highlighting their strengths and limitations.
    \item \textbf{Section 3} discusses approximation algorithms, which trade off solution quality for improved computational efficiency.
    \item \textbf{Section 4} delves into heuristic methods, both constructive and improvement-based, showcasing how these methods build and refine solutions iteratively.
    \item \textbf{Section 5} introduces metaheuristics, a class of advanced heuristics inspired by natural processes and optimization theory, which are often used to tackle large-scale CVRP instances.
    \item \textbf{Section 6} focuses on Neural Combinatorial Optimization, detailing the methodologies, architectures, and learning paradigms used to apply NCO to CVRP. This section also covers experimental evaluations of NCO techniques, including supervised learning methods, reinforcement learning frameworks like POMO, and techniques for improving model performance using instance augmentation and beam search.
\end{itemize}

By integrating traditional optimization techniques with cutting-edge neural models, this report provides a comprehensive exploration of the CVRP and its modern-day solutions. Our study demonstrates the potential of NCO as a transformative approach to solving combinatorial optimization problems, bridging the gap between theoretical research and practical applications.

%% file: exact.tex
\section{Exact Algorithms}
We briefly review the most recent works proposing exact algorithms for the CVRP; all of them are based on the combination of column and cut generation. 
\begin{enumerate}
    \item \textbf{Fukasawa et al. \cite{fukasawa2006robust}} presented a BCP algorithm having the following features:
    \begin{itemize}
        \item The columns are associated with the q-routes without k-cycles, a relaxation of the elementary routes that allow multiple visits to a customer, on the condition that at least k other costumers are visited between successive visits. The separated cuts are the same used in previous Branch-and-Cut algorithms over the two-index CVRP formulation. Those cuts are robust with respect to q-route pricing.
        \item If the column generation at the root node is found to be too slow, the algorithm automatically switches to a Branch-and-Cut.
    \end{itemize}
    All benchmark instances from the literature with up to 135 vertices could be solved to optimality, a significant improvement over previous methods.
    \item \textbf{Baldacci, Christofides, and Mingozzi \cite{baldacci2008exact}} presented an algorithm based on column and cut generation with the following features:
    \begin{itemize}
        \item The columns are associated with elementary routes. Besides cuts for the two-index formulation, Strengthened Capacity Cuts and Clique Cuts are separated. Those latter cuts are effective but non-robust; they make the pricing harder.
        \item A sequence of cheaper lower bounding procedures produces good estimates of the optimal dual variable values. The potentially expensive pricing is only called in the last stage, with the dual variables bounded to be above the estimates, obtaining convergence (to a bound slightly below the theoretical optimal) in fewer iterations.
        \item Instead of branching, the algorithm finishes in the root node (therefore, it is not a BCP) by enumerating all elementary routes with reduced cost smaller than the duality gap. A set-partitioning problem containing all those routes is then given to a Mixed-Integer Programming (MIP) solver.
    \end{itemize}
    The algorithm could solve almost all instances solved in \cite{fukasawa2006robust}, usually taking much less time. However, the exponential nature of some algorithmic elements, in particular the route enumeration, made it fail on some instances with many customers per vehicle.
    \item \textbf{Pessoa, Poggi de Aragao, and Uchoa \cite{pessoa2008robust}} presented some improvements over Fukasawa et al. \cite{fukasawa2006robust}:
    \begin{itemize}
        \item Cuts from an extended formulation with capacity indices were also separated. Those cuts do not change the complexity of the pricing of q-routes by dynamic programming.
        \item The idea of performing elementary route enumeration and MIP solving to finish a node was borrowed from \cite{baldacci2008exact}. However, in order to avoid a premature failure when the root gap is too large, it was hybridized with traditional branching.
    \end{itemize}
    \item \textbf{Baldacci, Mingozzi, and Roberti \cite{baldacci2011new}} improves upon Baldacci, Christofides, and Mingozzi \cite{baldacci2008exact} by the introduction of the following elements:
    \begin{itemize}
        \item The \textit{ng}-routes, a newly proposed relaxation that is more effective than the q-routes without \textit{k}-cycles, are used in the earlier bounding procedures and also to accelerate the pricing/enumeration of elementary routes. This latter part of the algorithm is also enhanced by considering multiple dual solutions. A route having reduced cost larger, with respect to any dual solution, than the current duality gap can be discarded.
        \item Subset Row Cuts and Weak Subset Row Cuts, which have less impact on the pricing with respect to Clique Cuts, are separated.
    \end{itemize}
    The resulting algorithm is not only faster on average, but it is much more stable than the algorithm in \cite{baldacci2008exact}, being able to solve even some instances with many customers per vehicle.
    \item \textbf{Contardo \cite{contardo2012new}} introduced new twists on the use of non-robust cuts and on route enumeration:
    \begin{itemize}
        \item The columns are associated with q-routes without 2-cycles, a relatively poor relaxation. The partial elementarity of the routes is enforced by non-robust Strong Degree Cuts. Robust cuts from two-index formulation and non-robust Strengthened Capacity Cuts and Subset Row Cuts are also separated.
        \item The enumeration of elementary routes is directed to a pool of columns. As soon as the duality gap is sufficiently small to produce a pool with reasonable size (a few million routes), the pricing starts to be performed by inspection. From this point, an aggressive separation of non-robust cuts can be performed, leading to very small gaps.
    \end{itemize}
    The reported computational results are very consistent. In particular, a hard in- stance from the literature, M-n151-k12 (151 vertices, 12 vehicles), was solved to optimality, setting a new record.
    \item \textbf{Ro{}pke \cite{ropke2012branching}} went back to robust BCP. The main differences from \cite{fukasawa2006robust} are the following:
    \begin{itemize}
        \item Instead of \textit{q}-routes without \textit{k}-cycles, the more effective \textit{ng}-routes are used.
        \item A sophisticated and aggressive strong branching is performed, drastically reducing the average size of the enumeration trees.
    \end{itemize}
    The overall results are comparable with the results in \cite{contardo2012new} and \cite{baldacci2011new}. A long run of that algorithm could also solve M-n151-k12.
    \item \textbf{Contardo and Martinelli \cite{contardo2014new}} improved upon Contardo \cite{contardo2012new}:
    \begin{itemize}
        \item Instead of \textit{q}-routes without 2-cycles, \textit{ng}-routes are used. Moreover, the performance of the dynamic programming pricing was enhanced by the use of the DSSR technique \cite{righini2008new}.
        \item Edge variables are fixed by reduced costs, using the procedure proposed in \cite{irnich2010path}.
    \end{itemize}
    \item Finally, \textbf{Pecin et al. \cite{pecin2017improved}} proposed a BCP that incorporates elements from all the previous algorithms, usually enhanced and combined with new elements:
    \begin{itemize}
        \item The most important original contribution is the introduction of the limited memory Subset Row Cuts. They are a weakening of the traditional Subset Row Cuts that can be dynamically adjusted, making them much less costly in the pricing, and yet without compromising their effectiveness.
        \item It uses capacity indices \cite{pessoa2008robust} in its underlying formulation. This allows a fixing of variables by reduced costs that is superior to fixing in \cite{irnich2010path}.
        \item The columns in the BCP are associated with \textit{ng}-routes. The dynamic programming pricing uses bidirectional search that differs a little from that proposed in \cite{righini2006symmetry} because the concatenation phase is not necessarily performed at half of the capacity. A column generation stabilization by dual smoothing \cite{pessoa2013out} may also be employed.
        \item The BCP hybridizes branching with route enumeration. Actually, it performs an aggressive hierarchical strong branching.
        \item When the addition of a round of non-robust cuts makes the pricing too slow, the BCP performs a rollback. The offending cuts are removed even if the lower bound of the node decreases.
    \end{itemize}
    The algorithm could solve all the classical instances from the literature with up to 200 vertices in reasonable times, including the hard instances M-n200-k17 and M-n200-k16.
\end{enumerate}

%% file: approximation.tex
\section{Approximation Algorithms}

For general metrics, Haimovich and Kan \cite{haimovich1985bounds} considered a simple heuristic, called tour partitioning, which starts from a TSP tour and then splits it into tours of size at most $Q$ by making back-and-forth trips to $r$ at certain points along the TSP tour. 
They showed this gives a $(1 + (1 - 1/Q)\alpha)$-approximation for splittable CVRP, where $\alpha$ is the approximation ratio for TSP. Essentially the same algorithm yields a $(2 + (1 - 2/Q)\alpha)$-approximation for unsplittable CVRP \cite{altinkemer1987heuristics}.
These stood as the best-known bounds until recently, when Blauth et al. \cite{blauth2023improving} showed that given a TSP approximation $\alpha$, there is an $\epsilon > 0$ such that there is an $(\alpha + 2 \cdot (1 - \epsilon))$-approximation algorithm for CVRP. For $\alpha = 3/2$, they showed $\epsilon > 1/3000$. 
They also describe a $(\alpha + 1 - \epsilon)$-approximation algorithm for unit demand CVRP and splittable CVRP. 
Friggstad et al. \cite{friggstad2022improved} presented two approximation algorithms for unsplittable CVRP: a combinatorial $(\alpha + 1.75)$-approximation, where $\alpha$ is the approximation factor for the Traveling Salesman Problem, and an approximation algorithm based on LP rounding with approximation guarantee $\alpha + ln(2) + \delta \approx 3.194 + \delta$ in $n^{O(1/\delta)}$ time. Both approximations can further be improved by a small amount when combined with recent work by Blauth, Traub, and Vygen (2021). 

Approximation by Friggstad et al. is the current best for the general case. However, there have been improvements for special cases in recent years. CVRP remains APX-hard in general metrics in this case but is polynomial-time solvable on trees. There exists a PTAS for CVRP
in the Euclidean plane as shown by Khachay et al. \cite{khachay2016ptas}. A PTAS for planar graphs was given by Becker et al. \cite{becker2019ptas} and a QPTAS for planar and bounded-genus graphs was then given by Becker et al. \cite{becker2017quasi}. A PTAS for graphs of bounded highway dimension and an exact algorithm for graphs with treewidth $tw$ with running time $O(n^{tw\\cdot Q})$ was shown by Becker et al. \cite{becker2018polynomial}. Cohen-Addad et al. \cite{cohen2020light} showed an efficient PTAS for graphs of bounded-treewidth, an efficient PTAS for bounded highway dimension, an efficient PTAS for bounded genus metrics and a QPTAS for minor-free metrics.

%% file: heuristics.tex
\section{Heuristics}
Exact methods have a guarantee of solution optimality and are more the-
oretically sound. However, it is difficult for them to handle the increasingly complex modern real-world vehicle routing problems efficiently. 
In contrast, heuristics provide a high-quality solution
at a reasonable computational cost and have good generalization ability, and thus have gained popularity among both researchers and practitioners. The heuristics are categorized into two main categories: 1) Constructive heuristics, 2) Improvement Heuristics
\subsection{Constructive Heuristics}
Constructive heuristics construct routing solutions from zero following
some fixed empirical procedures. They build a solution quickly, yet there
is usually a gap between its result and the optimal one. The existing con-
structive heuristics can be summarized into four algorithm frameworks: 1)
nearest neighbor method, 2) insert method, 3) saving method, and 4) sweep
method.
\subsubsection{Nearest Neighbor Method}
The routes can be built either sequentially or parallelly. In sequential route building, a route is extended by greedily adding the nearest feasible unrouted customer with the depot as the starting node. A new route is initialized from the depot when no customer can be added. The shortcoming of sequential route building is that the last vehicle usually has a lower loading ratio compared to other routes. The shortcoming can be alleviated if the routes are constructed in parallel. In parallel route building, the number of vehicles $K$ is set beforehand and the $K$ routes are extended in parallel. In each iteration, each route is extended with the closest unrouted customer, which means $K$ customers will be added. The process is iteratively performed until all the customers are visited. If customers can not be added to the $K$ routes, then a new route will be created following the sequential route-building strategy. The time complexity of the nearest neighbor method with n customers is $O(n^2)$, The algorithm iterates n times and takes $O(n)$ to find the nearest neighbor in each iteration.

\subsubsection{Insert Method}
It initializes some empty routes and inserts the unrouted customers one by one into the routes but not necessarily into the end of each route, which is the solution in the nearest neighbor method. In each iteration, the insertion that has the minimum cost is performed. The worst time complexity
of the insert method is $O(n^3)$. Similar to the nearest neighbor method, the insert method iterates $n$ times. However, there are two layers of minimization problem in each iteration: first, to find the minimal insertion cost for each unrouted customer and then find the unrouted customer with the cheapest minimal insertion cost.

\subsubsection{Saving Method}
It starts with an initial solution, in which a different
route serves each customer, and iteratively combines the short routes into longer routes with lower overall costs. 
The procedure can be performed in parallel or in sequence. 
In its parallel version, the method iteratively combines two routes with the endpoints $i$ and $j$, which produces the maximum feasible distance saving $s_{ij} = c_{i0} + c_{0j} - c_{ij}$. 
In the sequential version, each route is considered in turn. One route is iteratively extended by feasible saving operation until no such feasible merge can be used. The time complexity of a straightforward implementation of the saving method is $O(n^3)$. The algorithm starts with $n$ routes. 
It takes $n$ merge operations (iterations) and each merge operation is decided at the cost of $O(n^2)$ if every maximum feasible distance saving $s_{ij} = c_{i0} + c_{0j} - c_{ij}$ is checked or rechecked in each iteration. To reduce the complexity, the $n(n - 1)/2$ combinations corresponding to $n(n - 1)/2$ possible mergers can be computed and sorted at the beginning of the algorithm. The sort can be finished in $O(n^2log_2n)$ using a fast sorting algorithm. 
\subsubsection{Sweep Method}
The algorithm first sets the depot as the origin and sorts the nodes according to the polar angle. In its basic version, the node is added to the route circularly. A new route will be created if the insertion is infeasible. Another approach is a cluster-first, route-second method. All
the customers are clustered into several clusters according to the polar angle and a TSP is solved in each cluster. The performance of the sweep method can be greatly affected by the location of the depot. A poor result will be generated when the depot is off-centered. In addition, the last route often has a lower loading ratio because the clusters are built one by one. The time complexity of the basic sweep method is $O(n)$ because all the required computing is adding the next unrouted customer to the previous route according to the polar angle. In more advanced versions, the time complexity mainly depends on the method used for the TSPs.

\subsection{Improvement Heuristics}
Improvement heuristics explore the neighborhood of the incumbent so-
lution to achieve an objective improvement. They can quickly converge to
the local optimal and thus are efficient at solving large-scale routing problems. There are two widely recognized classes of improvement heuristics:
intra-route and inter-route. The difference between the two classes is the
neighborhood structure. The former searches inside a single route, while the
latter involves multiple routes.
\subsubsection{Intra-route Method}
Intra-route improvement heuristics explore the neighborhood involving
only one route. Most of them originated from the local search operators
for TSP. For instance, among the simplest ones, one customer can be
relocated to a different position in the same route, or two customers in one
route can be exchanged. 
The descriptions for some heuristics are given as follows:
\begin{itemize}
    \item \textit{Relocate}: Relocate selects one route from the solution and relocates one node to another position in the route. The time complexity of finding the best relocation solution is $O(n^2)$.
    \item \textit{Exchange}: Exchange selects one route and exchanges the positions of two nodes in the route. The time complexity of finding the best relocation solution is $O(n^2)$.
    \item \textit{2-opt,3-opt,$\lambda$-opt}: 2-opt first selects one route and then reconnects the ends of two edges in the route. In other words, it reverses a sequence of nodes within the route. The time complexity of finding the best reconnection is $O(n^2)$. The 2-opt can be regarded as a subset of 3-opt, which reconnects three edges with a time complexity of $O(n^3)$. $\lambda$-opt further generalizes them to considering $\lambda$ edges.
    \item \textit{Or-exchange}: Or-exchange selects a sequence of nodes with a preset length and relocates it to another position in the same route. The or-exchange can be regarded as a subset of 3-opt. Different from 3-opt, it only needs to specify the location of the start (or end) node of the sequence and the corresponding destination in the route. Therefore, the time complexity of finding the best exchange is $O(n^2)$.
    \item \textit{GENI}: GENI selects one route and a subset of three vertices. For a vertex $v$ not yet on the route, it implements the least cost insertion considering the two possible orientations of the tour and the two insert types. The time complexity of each insertion is $O(np^4 + n^2)$ when the neighborhood size is $p$.
\end{itemize}

\subsubsection{Inter-route Method}
Inter-route heuristics involve local searches across multiple routes. Many
of them are extensions of intra-route counterparts. For example, insert and
swap are the extension of relocate and exchange, respectively. The former
removes a customer from one route and reinserts it into another route. The
latter swaps two customers from different routes. Some representative inter-route improvement heuristics are introduced as follows:

\begin{itemize}
    \item \textit{2-opt*}: 2-opt* selects two routes and reconnects the end of two edges [218]. The reconnection with the best cost reduction is chosen. If the cost of the selection of two routes can be neglected, its time complexity is the same as 2-opt.
    \item  \textit{Insert}: Insert selects one node and inserts it into another position. Similar to the Relocation operator, the time complexity is $O(n^2)$.
    \item \textit{Swap}: Swap selects two nodes and swaps their locations [210]. Like the Exchange operator, the time complexity is $O(n^2)$. 
    \item \textit{CROSS}: CROSS exchanges two customer sequences (one of the two sequences can be empty). It generalizes the three mentioned operators: 2-opt*, Insert, and Swap. The time complexity is at most $O(n^4)$ and $O(\lambda^2n^2)$ when the size of the sequence is limited by a value $\lambda$.
    \item \textit{$\lambda$-interchange}: $\lambda$-interchange further generalizes CROSS. It allows exchanging any set of less than $\lambda$ nodes between two routes. The set can be non-consecutive, empty, and reversed during the reinsert.
\end{itemize}

%% file: metaheuristics.tex
\section{Metaheuristics}
Metaheuristics are presented in a more general and high-level way, in contrast to constructive heuristics and improvement heuristics, which are problem-dependent and attempt to exploit the feature and structure of the
target problem. According to the population management strategy, we classify metaheuristics into two categories: 1) single-solution-based and 2) population-based method. The widely-used metaheuristics in vehicle routing includes simulated annealing (SA), tabu search (TS), iterated local search (ILS), large neighborhood search (LNS), genetic algorithm (GA), ant colony optimization (ACO), memetic algorithm (MA). These algorithms are among the top-used algorithms in vehicle routing as reviewed in \cite{elshaer2020taxonomic}. Among them, GA received the most studies, followed by TS, LNS, and ACO. The number of publications on LNS and MA is growing rapidly in the last decade. Undoubtedly, these algorithms are very efficient in solving any complex optimization problem but they also come with the constraints of slow or premature convergence. Generally, to overcome these constraints, hybrid methods are used. Numerous metaheuristic approaches have been used to solve CVRP, which despite their modest computational capacity, have been proven to be quite effective. 

\begin{enumerate}

\item \textbf{Prins \cite{prins2004simple}} was the first to propose a hybrid genetic algorithm capable of solving vehicle routing problem which was tested on 34 benchmark instances. Notably, prior to the emergence of this hybrid genetic algorithm, genetic algorithms had earlier been utilized to address the CVRP. However, its performance consistently lagged behind compared to other metaheuristic methods. The hybridization of GA with a local search technique by Prins improved the solution accuracy and prevented it from premature convergence. The hybrid GA suffers from slow convergence speed. 

\item \textbf{Nazif \& Lee \cite{nazif2012optimised}} proposed an optimized crossover operator for solving CVRP utilizing genetic algorithm. It was examined on 27 benchmark instances. 

\item \textbf{Teoh and his collaborators \cite{teoh2015differential}} developed a local search-based improved differential evolution technique (DELS) which was experimented on 88 well-known benchmark instances. Three different local search moves were employed: swapping, dropping one point, and flipping. This implementation improved the exploration and exploitation capability of DELS. 

\item \textbf{Hosseinabadi et al. \cite{hosseinabadi2017new}} formulated a new CVRP-solving method based on gravity emulation local search (CVRP\_GELS). The CVRP\_GELS algorithm was designed utilizing two of the four basic parameters of gravitational force and velocity in physics based on the concepts of random search and searching agents, which is a collection of masses interacting with each other according to newtonian gravity and the laws of motion. This method was examined on 175 benchmark instances. One of the merits of this algorithm is its low execution time. The algorithm's performance was less satisfactory for small-scale instances but proved highly effective for larger-scale ones. 

\item \textbf{Faiz et al. \cite{faiz2018efficient}} devised a mechanism known as PVNS-ASM to handle capacitated vehicle routing problem by integrating perturbation-based variable neighborhood search with an adaptive selection approach to increase population diversity, enhance exploration and exploitation capabilities, and avoid premature convergence to locally optimal solutions. It was examined on 21 benchmark instances. 

\item \textbf{Sbai et al. \cite{sbai2022two}} constructed a hybrid metaheuristic HGA-VNS. This hybrid approach combines the strengths of genetic algorithm in exploration and the potent exploitation capabilities of variable neighborhood search (VNS). The algorithm outperformed majority of the state-of-the-art methods after a series of tests that involved roughly around 186 instances. 

\item \textbf{Altabeeb et al. \cite{altabeeb2019improved}} put forth a hybrid firefly algorithm termed as CVRP-FA. To increase the quality of the solution and hasten convergence, they combined 2h-opt and improved 2-opt algorithms. Additionally, partially mapped crossover as well as two mutation operators were also employed. The effectiveness of this technique was examined on 82 well-known benchmark instances. 

\item \textbf{Sajid et al. \cite{sajid2021novel}} developed an NSGA-II-based routing algorithm using a new tour best cost crossover operator to solve bi-objective CVRP. The algorithm aimed at improving solution quality, avoiding being entrapped in local optima, and accelerating convergence speed. It was evaluated on 88 benchmark instances. 

\item \textbf{Jiang et al. \cite{jiang2022evolutionary}} designed a relevance matrix-based evolutionary algorithm, termed as RMEA, to solve CVRP. To evolve the population, a relevance matrix-based crossover operator was designed, and a relevance matrix-based diversity preservation strategy was put forth as a method to increase population diversity. The algorithm was examined on 37 benchmark instances. 

\item \textbf{Khoo and Mohammad \cite{khoo2021parallelization}} have also devised a hybrid ruin \& recreated genetic algorithm HRRGA to solve multi-objective vehicle routing problem with time windows.

\item \textbf{Vidal et al. \cite{vidal2022hybrid}} developed HGS-CVRP. Hybrid Genetic Search (HGS) combines elements of genetic algorithms (a global search strategy) with local search techniques to refine solutions further 

\item Similarly, other researchers have also devised different techniques for solving capacitated vehicle routing problem.

\end{enumerate}

%% file: nco.tex
\section{Neural Combinatorial Optimization}

Neural Combinatorial Optimization (NCO) is an emerging framework that leverages deep learning to solve complex combinatorial optimization problems. Unlike traditional optimization methods, NCO integrates neural networks to learn representations and decision-making strategies directly from data. This paradigm enables generalization across problem instances, scalability to large problem sizes, and adaptability to varied problem structures. Based on the deep learning paradigm, NCO can broadly be categorized into two approaches: supervised learning and reinforcement learning.

\subsection{Overview}
\label{sec:overview}

In supervised learning, NCO models are trained on labeled datasets consisting of problem instances and their corresponding optimal solutions. Let \( x \in \mathbb{R}^d \) denote the input features of a combinatorial problem instance and \( y \in \mathbb{R}^k \) represent the optimal solution. The goal is to learn a function \( f_\theta(x) \) parameterized by a neural network \( \theta \), such that \( f_\theta(x) \approx y \). The training process minimizes a loss function \( L(y, f_\theta(x)) \), typically the mean squared error or cross-entropy loss. For instance, in the Capacitated Vehicle Routing Problem (CVRP), \( x \) could encode customer locations and demands, and \( y \) would represent the optimal route. A common architecture involves an encoder-decoder framework, where the encoder generates a fixed-size embedding for \( x \), and the decoder sequentially predicts components of \( y \). \\

Reinforcement learning-based NCO methods aim to maximize a reward signal \( R(y) \) that quantifies the quality of the solution \( y \). Here, the model interacts with an environment representing the problem, taking actions \( a_t \) at each step \( t \) to construct \( y \). The probability of taking an action is governed by a policy \( \pi_\theta(a_t | s_t) \), where \( s_t \) is the state at time \( t \). The objective is to optimize \( \theta \) to maximize the expected reward:
\[
J(\theta) = \mathbb{E}_{\pi_\theta} \left[ \sum_{t=1}^T R(a_t) \right].
\]
Policy gradient methods, such as REINFORCE, are commonly used to update \( \theta \) via the gradient:
\[
\nabla_\theta J(\theta) = \mathbb{E}_{\pi_\theta} \left[ R(y) \nabla_\theta \log \pi_\theta(y) \right].
\]
Techniques like Policy Optimization with Multiple Optima (POMO) enhance the reinforcement learning framework by reducing biases from initial states and enabling efficient exploration of multiple optima.

\subsection{Supervised Learning}
\subsubsection{Literature Review}
\label{sec:literature_review}

Supervised learning approaches in Neural Combinatorial Optimization (NCO) aim to train models using labeled datasets containing input-output pairs of problem instances and their optimal solutions. These methods have been pivotal in demonstrating the ability of neural networks to generalize and approximate high-quality solutions for combinatorial problems.

Luo et al. (2023) proposed the Heavy Decoder architecture, a supervised learning framework specifically designed for routing problems like the Capacitated Vehicle Routing Problem (CVRP)\cite{luo2023neural}. Their model employs a lightweight encoder that processes input features and a computationally intensive decoder that generates solutions sequentially. The heavy decoder leverages multi-head attention mechanisms to capture dependencies between customer nodes, significantly enhancing solution quality. This architecture achieved state-of-the-art performance on large-scale CVRP instances, demonstrating strong generalization across problem sizes.



Supervised learning provides a strong foundation for NCO by utilizing optimal solutions to guide model training. These methods have proven particularly effective for problems with structured datasets, offering a pathway to scalable and high-performance combinatorial optimization solutions.

\subsubsection{Light Encoder and Heavy Decoder}
\begin{figure}[t]
  \centering
  \includegraphics[width=0.99\linewidth]{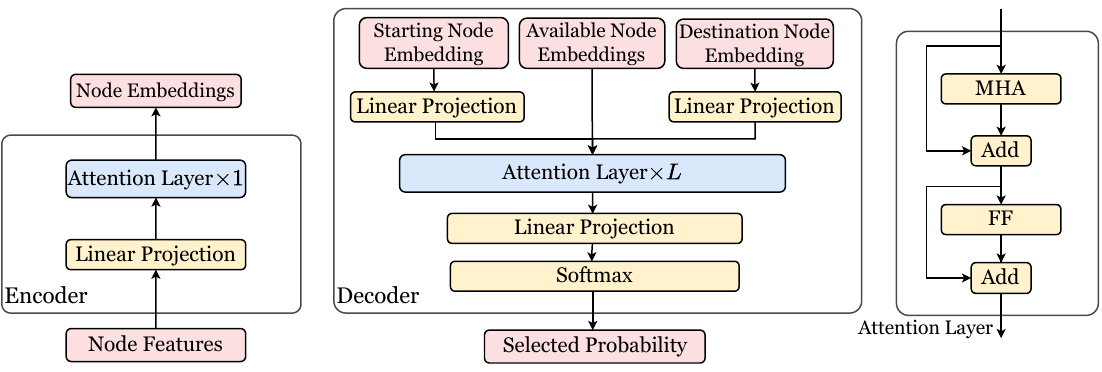}
  \caption{The structure of our proposed LEHD model, which has a single-layer light encoder and a heavy decoder with $L$ attention layers.}
  \label{LEHD model architecture}
  \vspace{-1em}
\end{figure}

We introduce a novel NCO model with the Light Encoder and Heavy Decoder (LEHD) structure to tackle the critical issue regarding large-scale generalization. The LEHD-based model consists of a light encoder and a heavy decoder, as shown in Figure \ref{LEHD model architecture}. The light encoder has one attention layer, and the heavy decoder has L attention layers.

\paragraph{Encoder}
For a problem instance $\mathbf{S}$ with $n$ node features $(\mathbf{s}_1,\ldots,\mathbf{s}_n)$ (e.g., the coordinates of $n$ cities), a constructive model parameterized by $\boldsymbol{\theta}$ generates the solution in an autoregressive manner, i.e., it constructs the solution by selecting nodes one by one. The light encoder transforms each node features $\mathbf{s}_i$ to its initial embedding $\mathbf{h}_i^{(0)}$ through a linear projection. The initial embeddings $\{\mathbf{h}_1^{(0)},\ldots,\mathbf{h}_n^{(0)}\}$ are then fed into one attention layer to get the node embedding matrix $H^{(1)}=(\mathbf{h}^{(1)}_1,\ldots,\mathbf{h}^{(1)}_n)$.

\paragraph{Attention Layer}
The attention layer comprises two sub-layers: the multi-head attention (MHA) sub-layer and the feed-forward (FF) sub-layer \citep{vaswani2017attention}. Different from generic NCO models such as AM \citep{kool2018attention}, the normalization is removed from our model to enhance generalization performance (see Appendix~\ref{Ablation study of normalization}). Let $H^{(l-1)}=(\mathbf{h}_1^{(l-1)},\ldots,\mathbf{h}_n^{(l-1)})$ be the input of the $l$-th attention layer for $l =1,\ldots, L$, the output of the attention layer in terms of the $i$-th node is calculated as: 
\begin{equation}
\begin{aligned}
    \hat{\mathbf{h}}_i^{(l)} &= \mathbf{h}_i^{(l-1)} + \operatorname{MHA}(\mathbf{h}_i^{(l-1)},H^{(l-1)}),\\
    \mathbf{h}_i^{(l)} &= \mathbf{\Hat{h}}_i^{(l)} + \text{FF}(\mathbf{\Hat{h}}_i^{(l)}).
\end{aligned}
\end{equation}
We denote this embedding process as $H^{(l)} = \operatorname{AttentionLayer}(H^{(l-1)})$.

\paragraph{Decoder}
The decoder sequentially constructs the solution in $n$ steps by selecting node by node. In the $t$-th step for $t\in \{1,\ldots,n\}$, the current partial solution can be written as $(x_1,\ldots, x_{t-1})$, the first selected node's embedding is denoted as $\mathbf{h}^{(1)}_{x_1}$, the embedding of the node selected in the previous step is denoted as $\mathbf{h}^{(1)}_{x_{t-1}}$, and the embeddings of all available nodes is denoted by $H_a =\{\mathbf{h}^{(1)}_i \mid i \in \{1,\ldots,n\}\setminus\{x_{1},\ldots,x_{t-1}\} \}$. Since the decoder has $L$ attention layers, the $t$-th construction step of LEHD can be described as:
\begin{equation}
\label{LEHD decoder equation}
\begin{aligned}
       \widetilde{H}^{(0)} &= \operatorname{Concat}(W_1\mathbf{h}^{(1)}_{{x_1}},W_2\mathbf{h}^{(1)}_{{x_{t-1}}},H_a),\\
        \widetilde{H}^{(1)} & = \operatorname{AttentionLayer}(\widetilde{H}^{(0)}),\\
        & \cdots \\
        \widetilde{H}^{(L)} & = \operatorname{AttentionLayer}(\widetilde{H}^{(L-1)}),\\
        u_i &= \begin{cases}
           W_O\widetilde{\mathbf{h}}_{i}^{(L)}, &\text{$i\neq1$ or $2$}\\
           -\infty, &\text{otherwise}
           \end{cases},\\
        \mathbf{p}^t &= \operatorname{softmax}(\mathbf{u}),
\end{aligned}
\end{equation}
where $W_1, W_2, W_O$ are learnable matrices. The matrices $W_1$ and $W_2$ are used to re-calculate the embeddings of the starting node (i.e., the node selected in the previous step) and the destination node (i.e., the node selected in the first step), respectively. The matrix $W_O$ is used to transform the relation-denoting embedding matrix $\widetilde{H}^{(L)}$ into a vector $\mathbf{u}$ for the purpose of calculating the selection probability $\mathbf{p}^t$. The most suitable node $x_t$ is selected based on $\mathbf{p}^t$ at each decoding step $t$. Finally, a complete solution $\mathbf{x}=(x_{1},\ldots,x_{n})^\intercal$ is constructed by calling the decoder $n$ times.

The heavy decoder dynamically re-embeds the embeddings of the starting node, destination node, and available nodes via $L$ attention layers, thus updating the relationships among the nodes at each decoding step. Such a dynamic learning strategy enables the model to adjust and refine its captured relationships between the starting/destination and available nodes. In addition, as the size of the nodes varies during the construction steps, the model tends to learn the scale-independent features. These good properties enable the model to dynamically capture the proper relationships between the nodes, thereby making more informed decisions in the node selection on problem instances of various sizes.

\subsubsection{Codebase}
The GitHub link for the official code is: \url{https://github.com/CIAM-Group/NCO_code}. We have modified the code to improve the performance. Github link: \url{https://github.com/mahirlabibdihan/NCO_code}. 
An overview of the main modules is given below:
\begin{itemize}
    \item VRPTrainer: Main training loop.
    \item VRPTester: Main testing loop.
    \item VRPModel: Encoder-Decoder Model Architecture.
    \item VRPEnv: Maintains state of a batch.
\end{itemize}

\subsubsection{Dataset}
The standard data generation procedure of the previous work \cite{kool2018attention} was followed to generate CVRP datasets. The training sets consist of one million instances, each with 100 nodes. Optimal solutions are obtained using HGS-CVRP \cite{vidal2022hybrid}. The test set includes 10,000 instances with 100 (CVRP100) nodes and 128 instances for each of 200 (CVRP200), 500 (CVRP500), and 1000 (CVRP1000) nodes. Optimal solutions are obtained using LKH3 \cite{helsgaun2017extension}. The training and test datasets can be downloaded from Google Drive\footnote{\url{https://drive.google.com/drive/folders/1LptBUGVxQlCZeWVxmCzUOf9WPlsqOROR?usp=sharing}} or Baidu Cloud\footnote{\url{https://pan.baidu.com/s/12uxjol_5pAlnm0j4F6D_RQ?pwd=rzja}}.


\subsubsection{Methodology}
During the inference phase, LEHD model employs a greedy step-by-step approach to construct the whole solution. The tour generated via greedy search could be not optimal and contain multiple suboptimal local segments (e.g., partial solutions). Therefore, rectifying these suboptimal segments during the inference phase can effectively enhance the overall solution quality. For this, Random Re-Construct(RRC) is proposed. RRC randomly samples a partial solution from the initial solution and restructures it to obtain a new partial solution. If the new partial solution is superior, it replaces the old one.

We have modified the RRC approach by incorporating Simulated Annealing (SA) when determining whether to accept a new partial solution. Instead of always replacing the old partial solution with a superior one, the acceptance of the new partial solution is guided by the SA mechanism. This introduces a probabilistic component, allowing for the acceptance of potentially inferior solutions early in the process to escape local optima.

Simulated Annealing evaluates the new partial solution based on a temperature parameter that decreases over time. At higher temperatures, the method is more likely to accept worse solutions, fostering exploration, while at lower temperatures, it becomes more selective, emphasizing exploitation of the best solutions. This modification improves the robustness of RRC and enhances the quality of the final solution by effectively refining suboptimal segments.

\subsubsection{Experiments}
In this section, we compare our method with LEHD model. We report the optimality gap. The optimality gap measures the difference between solutions achieved by different learning methods and the (near) optimal solutions obtained using LKH3. Note that, as LKH3 is not an exact algorithm, it will give sub-optimal solutions. So, the optimzality gap can be negative too.

The main experimental results on uniformly distributed CVRP instances are reported in Table \ref{tab:rrc-sa-comparison}. We have obtained results for 50, 100 and 200 iterations. The integration of Simulated Annealing has demonstrated better results in most cases.

\begin{table}[t]
\label{textual-stats-table}
\begin{center}
\small
\begin{tabular}{l|c c|c c|c c}
\hline
 &  \multicolumn{2}{c|}{\textbf{i50}} & \multicolumn{2}{c|}{\textbf{i100}} & \multicolumn{2}{c}{\textbf{i200}} \\
\hline
         & rrc & rrc+sa & rrc & rrc+sa & rrc & rrc+sa \\
CVRP100  & \textbf{\underline{0.535\%}} & 0.558\% & \textbf{\underline{0.272\%}} & 0.284\% & 0.106\% & \textbf{\underline{0.096\%}}\\
CVRP200  & 0.515\% & \textbf{\underline{0.440\%}} & 0.217\% & \textbf{\underline{0.169\%}} & -0.032\% & \textbf{\underline{-0.112\%}} \\
CVRP500  & 0.930\% & \textbf{\underline{0.891\%}} & \textbf{\underline{0.546\%}} & 0.596\% & 0.223\% & \textbf{\underline{0.189\%}} \\
CVRP1000 & 2.814\% & \textbf{\underline{2.705\%}} & 2.370\% & \textbf{\underline{2.252\%}} & \textbf{\underline{1.826\%}} & 1.860\%\\
\hline
Overall  & \textbf{\underline{0.568\%}} & 0.587\% & 0.300\% & \textbf{\underline{0.283\%}} & 0.127\% & \textbf{\underline{0.117\%}}\\ 
\hline
\end{tabular}
\caption{Experimental results with uniformly distributed instances. i50, i100, and i200 represent 50, 100, and 200 iterations, respectively.}
\label{tab:rrc-sa-comparison}
\end{center}
\end{table}

\subsection{Reinforcement Learning}
\subsubsection{Policy Gradient Framework}

Reinforcement learning focuses on solving sequential decision-making problems, where an agent interacts with an environment to maximize cumulative rewards.Here, we outline the key concepts, starting with the definition of a policy, the associated objective function, and the procedure to update the policy.

\paragraph{Policy Definition}

The policy \(\pi_\theta\) is a probabilistic mapping from the current state and problem instance to an action. Formally, it is represented as:
\[
\pi_\theta(a_{t+1} \mid \rho, \tau_t)
\]
where:
\begin{itemize}
    \item \(\theta\): Parameters of the policy that we aim to optimize.
    \item \(a_{t+1}\): The action to be taken at step \(t+1\).
    \item \(\rho\): The problem instance, sampled from a distribution of problem instances \(\mathcal{D}\).
    \item \(\tau_t = (x_1, \ldots, x_t)\): The solution instance, which includes all states traversed up to time \(t\).
\end{itemize}

The policy \(\pi_\theta\) represents the probability of taking action \(a_{t+1}\) at a given time \(t\), given the current problem instance and the solution constructed so far. This probability is critical for exploring and exploiting potential solutions.

\paragraph{Probability of a Complete Solution Path}

The probability of reaching any specific solution instance \(\tau\) (a sequence of states and actions) is the product of the probabilities of taking all individual steps:
\[
\pi_\theta(\tau) = \prod_t \pi_\theta(a_{t+1} \mid \rho, \tau_t).
\]
This formulation ensures that the overall probability of a trajectory is decomposed into stepwise decisions, allowing the optimization of the entire solution path.

\paragraph{Objective Function}
To train the policy, we define an objective function that quantifies the quality of the solution paths generated by the policy. The reward for a given solution instance \(\tau\) is represented by \(R(\tau)\). The objective is to maximize the expected reward over all possible trajectories and problem instances.

\[
J(\theta) = \mathbb{E}_{\rho \sim \mathcal{D}} \left[ \mathbb{E}_{\tau \sim \pi_{\theta, \rho}} \left[ R(\tau) \right] \right].
\]

Here:
\begin{itemize}
    \item \(R(\tau)\): The reward function evaluates how good a solution instance \(\tau\) is.
    \item \(\mathbb{E}_{\tau \sim \pi_{\theta, \rho}}\): The expectation over solution instances \(\tau\) sampled according to the policy \(\pi_\theta\) for a specific problem instance \(\rho\).
    \item \(\mathbb{E}_{\rho \sim \mathcal{D}}\): The expectation over the distribution of problem instances.
\end{itemize}

The goal is to optimize the policy parameters \(\theta\) such that the expected reward \(J(\theta)\) is maximized.

\paragraph{Policy Update}
To maximize the objective function, we use gradient ascent. The gradient of \(J(\theta)\) with respect to the policy parameters \(\theta\) guides how the policy should be updated. The update rule is:
\[
\theta \gets \theta + \alpha \nabla_\theta J(\theta),
\]
where:
\begin{itemize}
    \item \(\alpha\): The learning rate, which determines the step size of the updates.
    \item \(\nabla_\theta J(\theta)\): The gradient of the objective function with respect to the policy parameters \(\theta\).
\end{itemize}

This update rule ensures that the policy parameters are adjusted in the direction that increases the expected reward \(J(\theta)\). Over successive iterations, the policy becomes better at generating high-quality solution paths.

\subsubsection{POMO - Policy Optimization with Multiple Optima}

\paragraph{Motivation}
In combinatorial optimization (CO) problems, we often deal with tasks such as finding the shortest path, solving the Traveling Salesperson Problem (TSP), or similar graph-based problems. Here, we outline the key challenges associated with standard reinforcement learning methods and introduce the motivation for using POMO (Policy Optimization with Multiple Optimas) to mitigate these issues.

Assume we are given a combinatorial optimization problem with a group of nodes \(\{v_1, v_2, \ldots, v_M\}\) and a trainable policy network. The policy is tasked with generating a valid solution \(\tau\) step-by-step. Formally, the policy \(\pi_t\) for selecting the next action at step \(t\) is defined as:
\[
\pi_t =
\begin{cases}
    p_\theta(a_t \mid s) & \text{for } t = 1, \\
    p_\theta(a_t \mid s, a_{1:t-1}) & \text{for } t \in \{2, 3, \ldots, M\},
\end{cases}
\]
where:
\begin{itemize}
    \item \(a_t\): The action (node) selected at time step \(t\).
    \item \(s\): The state defined by the problem instance.
    \item \(p_\theta\): The policy parameterized by \(\theta\).
\end{itemize}

This framework enables the generation of solutions by incrementally selecting actions based on the current state and past actions.

In many CO problems, a single solution can often be expressed in multiple equivalent forms due to symmetry. For example, in the case of the Traveling Salesperson Problem (TSP):
\[
\tau = (v_1, v_2, v_3, v_4, v_5)
\]
is equivalent to:
\[
\tau' = (v_2, v_3, v_4, v_5, v_1),
\]
since both represent the same closed tour. However, standard learning methods often fail to exploit this equivalence and become heavily biased by the starting point of the solution.

The starting action significantly influences the agent's future decisions, leading to asymmetry in the learning process. For example:
\begin{itemize}
    \item If the agent starts at a specific node, it might consistently favor certain solution paths over others, even when they are equivalent.
    \item This bias reduces the diversity of explored solutions and limits the policy's ability to generalize across problem instances.
\end{itemize}

In order to solve this problem , POMO offers the following modifications: 

\begin{itemize}
    \item \textbf{Multiple Starting Points} \\ 
POMO begins with designating $N$ different nodes $\{a_1^1, a_2^1, \dots, a_N^1\}$ as starting points for exploration (Figure 2, (b)). The network samples $N$ solution trajectories $\{\tau_1, \tau_2, \dots, \tau_N\}$ via Monte-Carlo method, where each trajectory is defined as a sequence $\tau_i = (a_i^1, a_i^2, \dots, a_i^M)$ for $i = 1, 2, \dots, N$. \\
Previous methods use a fixed \textit{START} token , which induces bias based on the initial token. POMO reduces this bias by creating N solution trajectories from N different starting nodes. \\
Conceptually, explorations by POMO are analogous to
guiding a student to solve the same problem repeatedly from many different angles, exposing her to a
variety of problem-solving techniques that would otherwise be unused.

    \item \textbf{A shared baseline for policy gradients} \\
     Once we sample a set of solution trajectories $\{\tau_1, \tau_2, \dots, \tau_N\}$, we can calculate the return (or total reward) $R(\tau_i)$ of each solution $\tau_i$. To maximize the expected return $J$, we use gradient ascent with an approximation
$$
\nabla_\theta J(\theta) \approx \frac{1}{N} \sum_{i=1}^N (R(\tau_i) - b_i(s))\nabla_\theta \log p_\theta (\tau_i|s)
$$
where $p_\theta (\tau_i|s) \equiv \prod_{t=2}^M p_\theta (a_t|s, a_{1:t-1})$.
\\

 $b_i(s)$ is a baseline that one has some freedom of choice to reduce the variance of the sampled gradients. In principle, it can be a function of $a_1$, assigned differently for each trajectory $\tau_i$.
In POMO, however, we use the shared baseline,
$$
b_i(s) = b_{shared}(s) = \frac{1}{N} \sum_{j=1}^N R(\tau_j) \quad \text{for all } i.
$$ \\
The shared baseline used by POMO makes RL training highly resistant to
local minima.With the shared baseline, each trajectory
now competes with N - 1 other trajectories where no two trajectories can be identical.

    \item \textbf{Multiple greedy trajectories for inference} \\
    Neural network models designed for combinatorial optimization (CO) problems generally offer two modes for inference. It can use the greedy method or the sampling method. While sampled solutions typically yield lower rewards on average compared to the greedy solution, repeated sampling can uncover solutions with higher rewards at the cost of increased computational effort. \\

POMO introduces a novel approach by producing multiple deterministic greedy trajectories instead of relying on stochastic sampling. Starting from \(N\) different initial nodes \(\{a_{1}^{1}, a_{1}^{2}, \dots, a_{1}^{N}\}\), the model deterministically generates \(N\) distinct greedy trajectories. This allows for selecting the best solution among these trajectories, similar to the strategy in sampling mode. However, the \(N\) greedy trajectories from POMO generally outperform \(N\) sampled trajectories, offering better solutions. 

\end{itemize}

\subsubsection{Experiments}

Policy optimization involves selecting actions or decisions based on a trained policy during inference. Below are the techniques used for inference in this context, along with detailed descriptions:

\paragraph{Softmax}
\begin{itemize}
    \item The Softmax function converts raw output scores (logits) into a probability distribution over the possible next nodes.
    \item This allows probabilistic selection, with higher likelihood nodes receiving more weight.
    \item The formula for Softmax is:
    \[
    S(y)_i = \frac{\exp(y_i)}{\sum_{j=1}^n \exp(y_j)}
    \]
    where:
    \begin{itemize}
        \item \( y_i \): Logit value of the \(i\)-th node.
        \item \( n \): Total number of nodes.
    \end{itemize}
    \item This method maintains diversity in exploration by considering all nodes, weighted by their probabilities.
\end{itemize}

\paragraph{Greedy}
\begin{itemize}
    \item The Greedy method deterministically selects the node with the highest probability.
    \item It avoids randomness, directly choosing the node with the maximum value in the probability distribution.
    \item The next node is computed as:
    \[
    \text{Next node} = \arg\max_i(P_i)
    \]
    where:
    \begin{itemize}
        \item \( P_i \): Probability of the \(i\)-th node.
    \end{itemize}
    \item This method is computationally efficient but may lead to suboptimal solutions by limiting exploration.
\end{itemize}

\paragraph{Gumbel-Softmax Sampling}
\begin{itemize}
    \item This technique adds controlled random noise to the logits using the Gumbel distribution, promoting exploration.
    \item The modified probabilities are then passed through a Softmax function for selection.
    \item The formula for Gumbel-Softmax is:
    \[
    y_i = \frac{\exp\left(\frac{z_i + g_i}{\tau}\right)}{\sum_{j=1}^k \exp\left(\frac{z_j + g_j}{\tau}\right)}
    \]
    where:
    \begin{itemize}
        \item \( z_i \): Log-probability value for the \(i\)-th node.
        \item \( g_i \): Gumbel noise added to \(z_i\).
        \item \( \tau \): Temperature parameter controlling randomness.
        \item \( k \): Total number of nodes.
    \end{itemize}
    \item Lower values of \(\tau\) make the selection more deterministic, while higher values encourage exploration.
\end{itemize}
\begin{figure}[ht]
    \centering
    \includegraphics[width=0.7\textwidth]{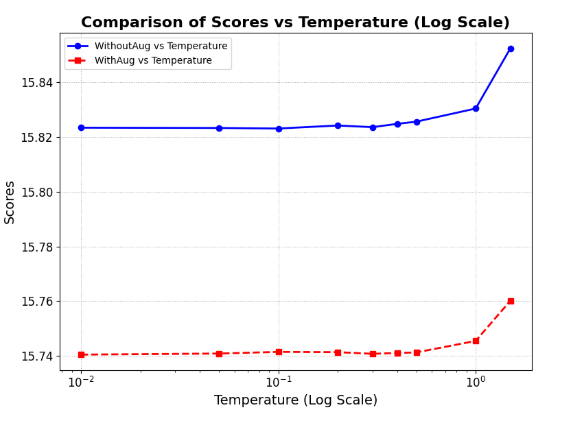} 
    \caption{Experiments using different temperatures}
\end{figure}

\paragraph{Epsilon-Greedy}
RL exploration strategy where the agent chooses a random action with probability \(\epsilon\) and the best-known action with probability \(1-\epsilon\).

\begin{figure}[ht]
    \centering
    \includegraphics[width=0.7\textwidth]{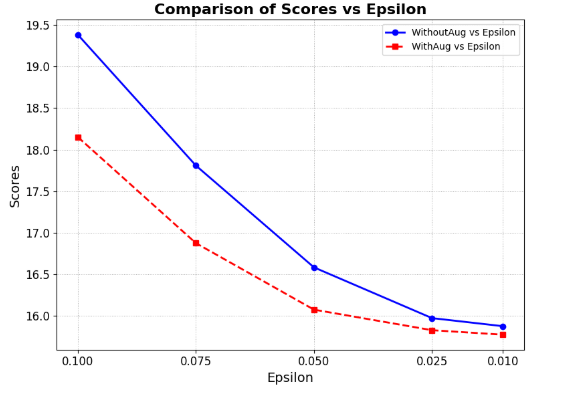} 
    \caption{Experiments using different values of epsilon}
\end{figure}
\[
a = 
\begin{cases} 
\text{random action}, & \text{with probability } \epsilon, \\
\text{best-known action}, & \text{with probability } 1 - \epsilon.
\end{cases}
\]

In summary, the results are shown in Table \ref{tab:decoding-strategy}.

\begin{table}[t]
\label{textual-stats-table}
\begin{center}
\small
\begin{tabular}{c|c|c}
\hline
\textbf{Method} & \textbf{Without Augmentation} & \textbf{With Augmentation} \\
\hline
\textbf{Argmax}                     & \textbf{\underline{15.8242}} & 15.7412 \\
\textbf{Softmax-Sampling}           & 15.8368 & 15.7544 \\
\textbf{Epsilon-Greedy}             & 15.8719 & 15.7695 \\
\textbf{Gumbel-Softmax}    & 15.8234 & \textbf{\underline{15.7405}} \\
\hline
\end{tabular}
\caption{Experiments using different Inference methods with and without
augmentation}
\label{tab:decoding-strategy}
\end{center}
\end{table}

\paragraph{Instance Augmentation}
One limitation of POMO's multi-greedy inference method lies in the fact that the number of greedy rollouts (\(N\)) it can perform is inherently constrained by the finite number of possible starting nodes. However, this limitation can be addressed by leveraging \textit{instance augmentation}, a natural extension of POMO's core idea. The objective of instance augmentation is to find alternative ways of reaching the same optimal solution. For instance, the problem can be reformulated so that the model perceives a different configuration but ultimately produces the same result. An example of this is flipping or rotating the coordinates of all nodes in a 2D routing optimization task, thereby creating new instances that allow for more diverse greedy rollouts.

In POMO, it uses 8 fold augmentation by default. 
\[
\begin{array}{cc}
(x, y) & (1 - x, y) \\
(x, 1 - y) & (1 - x, 1 - y) \\
(x, y) & (x, -y) \\
(-x, y) & (-x, -y)
\end{array}
\]
In our experiment , we modified those augmentations , to be \textbf{2-fold} , \textbf{4-fold} and \textbf{8-fold augmentation with rotations}. The result of the experiments is shown in Table \ref{tab:aug-strategy}.

\begin{table}[t]
\begin{adjustbox}{max width=\textwidth}
\small
\begin{tabular}{p{2cm}|p{2.7cm}|p{2.7cm}|p{2.7cm}|p{2.7cm}}
\hline
\textbf{Method} & \textbf{Augmentation 2 Fold} & \textbf{Augmentation 4 Fold} & \textbf{Augmentation 8 Fold (Horizontal and Vertical Flip)} & \textbf{Augmentation 8 Fold (Rotation)} \\
\hline
\textbf{Argmax}           & 15.7885             & 15.7915 & 15.7412 & 15.7743 \\
\hline
\textbf{Softmax-Sampling} & 15.8164             & 15.7586 & 15.7544 & 15.7676 \\
\hline
\textbf{Epsilon-Greedy}   & 15.7915             & 15.7908 & 15.7695 & 15.7992 \\
\hline
\textbf{Gumbel-Softmax}   & 15.7786    & 15.7582 & 15.7405 & 15.7678 \\
\hline
\textbf{Beam Search}   & \textbf{\underline{15.7422}}    & \textbf{\underline{15.7288}} & \textbf{\underline{15.6985}} & \textbf{\underline{15.7412}} \\
\hline
\end{tabular}
\end{adjustbox}
\caption{Experiments using different Augmentation methods}
\label{tab:aug-strategy}
\end{table}

\subsubsection{Adding Beam Search with POMO}
Our final technique that we used is to encorporate Beam search with POMO. Code is available at: \url{https://github.com/wjalal/pomo_beam}. 
\begin{itemize}
    \item Beam search is a heuristic search algorithm that explores a graph by expanding only a limited set of the most promising nodes, modifying best-first search to reduce memory usage.
    \item It uses breadth-first search to build a search tree. At each level, it generates all successors, sorts them by heuristic cost, and keeps only a fixed number ($\beta$, the beam width) of best state.
    \item Larger beam widths reduce pruning; infinite width makes it equivalent to best-first search. Beam width of 1 makes it identical to hill climbing.
    \item A goal state may be pruned, so it doesn't guarantee a solution if one exists, or the best solution. The beam width determines the memory required for the search.
\end{itemize}
In our approach , we maintain a beam\_size to indicate how the search for new nodes are explored. We also keep track of the log probability of generated trajectories in order to find the current best sequence. From $K$ initial states , beam search generates $K^2$ states. In the pruning state , we pick the top K states and continue.

\begin{table}[t]
\label{textual-stats-table}
\begin{center}
\small
\begin{tabular}{c|c|c}
\hline
\textbf{Method} & \textbf{Without Augmentation} & \textbf{With Augmentation} \\
\hline
\textbf{Argmax}                     & 15.8242 & 15.7412 \\
\textbf{Softmax-Sampling}           & 15.8368 & 15.7544 \\
\textbf{Epsilon-Greedy}             & 15.8719 & 15.7695 \\
\textbf{Gumbel-Softmax}             & 15.8234 & 15.7405 \\
\textbf{Beam Search}    & \textbf{\underline{15.7576}} & \textbf{\underline{15.6985}} \\
\hline
\end{tabular}
\caption{Experiments using different Inference methods with and without
augmentation}
\label{tab:decoding-strategy-beam}
\end{center}
\end{table}

\begin{algorithm}
\caption{Beam Search Forward}\label{alg:forward}
\begin{algorithmic}[1]
\Function{Forward}{state}
    \State $batch\_size \gets \text{state.BATCH\_IDX.size}(0)$
    \State $pomo\_size \gets \text{state.BATCH\_IDX.size}(1)$
    \State $beam\_size \gets \text{state.BATCH\_IDX.size}(2)$
    \State $prob \gets \mathbf{1}$ \Comment{Initialize probabilities to ones of size $(batch\_size, pomo\_size, beam\_size)$}
    \State $selected\_beams \gets \text{arange}(beam\_size)$ \Comment{Expand indices for all beams}
    \\
    \If{$state.selected\_count = 0$} \Comment{First move (depot)}
        \State $selected \gets \mathbf{0}$ \Comment{Zeros of size $(batch\_size, pomo\_size, beam\_size)$}\\
    \ElsIf{$state.selected\_count = 1$} \Comment{Second move (POMO)}
        \State $selected \gets \text{arange}(1, pomo\_size + 1)$ \Comment{Expand sequential indices for $beam\_size$}\\
    \ElsIf{$state.selected\_count = 2$} \Comment{Third move (BEAM)}
        \State $probs \gets \text{Decoder}(state.current\_node[:,:,0], state.load[:,:,0], state.ninf\_mask[:,:,0])$
        \State $(prob, selected) \gets \text{TopK}(probs, beam\_size)$\\
    \Else \Comment{Later moves}
        \For{$k \gets 0$ to $beam\_size-1$}
            \State $probs \gets \text{Decoder}(state.current\_node[:,:,k], state.load[:,:,k], state.ninf\_mask[:,:,k])$
            \State $(selected_k, prob_k) \gets \text{TopK}(probs, beam\_size)$
            \State $logprob_k \gets state.logprob[:,:,k] + \log(prob_k + 10^{-5})$
        \EndFor
        \State Concatenate results for $selected$, $prob$, and $logprob$
        \State Refine selections using top-k log probabilities
    \EndIf
    \State \Return $selected, selected\_beams, prob$
\EndFunction
\end{algorithmic}
\end{algorithm}

\begin{figure}[t]
  \centering
  \includegraphics[width=0.99\linewidth]{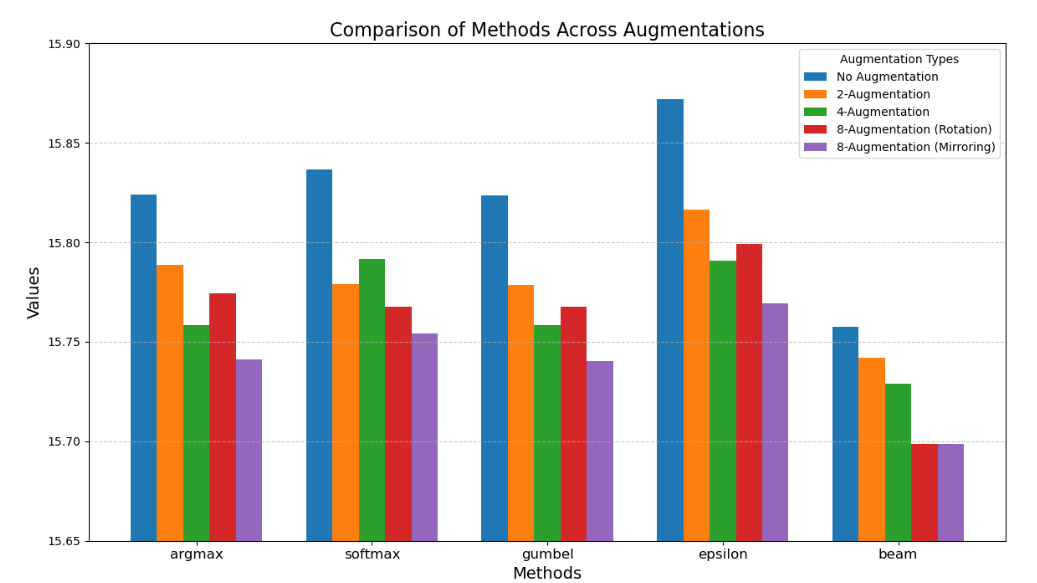}
  \caption{Comparison of different inference techniques and augmentation types.}
  \label{beam-results}
  \vspace{-1em}
\end{figure}